\documentclass[letterpaper]{article} 
\usepackage[]{aaai2026}  
\usepackage{times}  
\usepackage{helvet}  
\usepackage{courier}  
\usepackage[hyphens]{url}  
\usepackage{graphicx} 
\usepackage{natbib}  
\usepackage{caption} 
\usepackage{algorithm}
\usepackage{algorithmic}
\usepackage{booktabs}
\usepackage{amsmath}
\usepackage{multirow}
\usepackage{newfloat}
\usepackage{listings}
\DeclareCaptionStyle{ruled}{labelfont=normalfont,labelsep=colon,strut=off} 
\floatstyle{ruled}
\newfloat{listing}{tb}{lst}{}
\floatname{listing}{Listing}
\title{Sumudu Neural Operator for ODEs and PDEs}

\author { 
    Ben Zelenskiy\textsuperscript{\rm 1},
    Saibilila Abudukelimu\textsuperscript{\rm 1},
    George Flint\textsuperscript{\rm 2}\thanks{Senior authors.},
    Kevin Zhu\textsuperscript{\rm 1}\footnotemark[1],
    Sunishchal Dev\textsuperscript{\rm 1}\footnotemark[1]
}
\affiliations {
    \textsuperscript{\rm 1}Algoverse AI Research, USA\\
    ben.zelenskiy@gmail.com, saibi.kerim@gmail.com,\\
    kevin@algoverseacademy.com, dev@algoverseairesearch.org\\
    \textsuperscript{\rm 2}University of California, Berkeley, USA\\
    georgeflint@berkeley.edu
}

\begin{document}

\maketitle

\begin{abstract}
We introduce the Sumudu Neural Operator (SNO), a neural operator rooted in the properties of the Sumudu Transform. We leverage the relationship between the polynomial expansions of transform pairs to decompose the input space as coefficients, which are then transformed into the Sumudu Space, where the neural operator is parameterized. We evaluate the operator in ODEs (Duffing Oscillator, Lorenz System, and Driven Pendulum) and PDEs (Euler-Bernoulli Beam, Burger's Equation, Diffusion, Diffusion-Reaction, and Brusselator). SNO achieves superior performance to FNO on PDEs and demonstrates competitive accuracy with LNO on several PDE tasks, including the lowest error on the Euler-Bernoulli Beam and Diffusion Equation. Additionally, we apply zero-shot super-resolution to the PDE tasks to observe the model's capability of obtaining higher quality data from low-quality samples. These preliminary findings suggest promise for the Sumudu Transform as a neural operator design, particularly for certain classes of PDEs.\\
\textbf{Code:} \texttt{https://github.com/Lapuleu/SNO}
\end{abstract}

\section{Introduction}

\citet{Kovachki2021neural} introduces transform based neural operators, machines that learn maps between function spaces of infinite-dimensions. \citet{Boulle2024mathematical} discusses this completely new branch of neural architecture, explaining how they often utilize integral transforms to approximate operators, thereby enabling the learning of complex function mappings. They are particularly useful for solving differential equations compared to standard mathematical solvers, which are time and resource-intensive. Neural operators are not strictly as accurate as mathematical solvers, due to them estimating instead of solving, but they can be significantly more efficient. Two notable neural operators are the Fourier Neural Operator (FNO) and the Laplace Neural Operator (LNO). \citet{li2021fourier} introduce the Fourier neural operator (FNO), which decomposes the input space using a Fourier transform using the Fast Fourier Transform algorithm \cite{CooleyTukey1965}, yielding accurate PDE solutions more efficiently than previous analytical methods. \citet{cao2023lno} introduces the Laplace neural operator (LNO), which leverages the pole-residue formulation of a function in Laplace Space to decompose the input space—building upon FNO by using a more general integral transform than the Fourier Transform, as well as adding more sophistication (in the form of the pole-residue formulation) in the signal processing step.

\subsection{Neural Operators}
Neural operators are neural networks that learn mappings between functional spaces, enabling more generalization than traditional neural networks. The goal of any neural operator is to learn the nonlinear map $\mathcal{P_\phi}:\mathcal{X}\rightarrow \mathcal{Y}$ where $\phi$ is the network parameters. In recent years, neural operators have gained popularity in the realm of differential equations, originally inspired by the universal approximation theorem of operators proposed by \citet{ChenChen1995}. A significant motivator for the development of neural operators is their property of variable discretization, which means that the output discretization can be increased without retraining the model, allowing for high versatility as well as the possibility of zero-shot super-resolution.

The first functional neural network, DeepONet \citet{Lu2021learning}, functionally was the sum of two or more deep neural networks (DNNs) where the output(s) would have a trunk neural network and the input(s) would have branch neural networks. The DNNs had general architectures, meaning the architectures could alter the behavior of the neural network \cite{Lu_Jin_Karniadakis_2020}. In 2020, the groundbreaking Fourier Neural Operator showed the capabilities of neural operators, demonstrating both zero-shot super-resolution as well as high performance on a myriad of PDE tasks \cite{li2021fourier}. A similar architecture is demonstrated in the Laplace Neural Operator (LNO); however, LNO leverages the Laplace Transform as well as a pole-residue formulation of the Laplace Space, which allowed them to solve the issue of transient responses and no damping conditions that FNO struggled with.
We propose to again replace the Fourier transform in FNO with the Sumudu transform, a related integral transform that is far less studied than either the Laplace or Fourier transforms. We are motivated by some key properties of the Sumudu transform that allow us to define it in terms of a polynomial approximation, which has both computation efficiency benefits and proposed higher accuracy on zero-shot super resolution tasks for time-dependent ODEs/PDEs.
\paragraph{Transient Responses and Dampening} A transient response is the initial behavior of a signal. The more underdamped a signal, the more oscillitations it takes to reach steady-state. A steady-state is an equilibrium at which transients no longer matter and the signal goes on infinitely. The most fundamental example is a pure sinusoidal wave. Capturing transient responses has been a challenge for certain neural operator architectures but was significantly improved on by \citet{cao2023lno}.

\section{Methods}
\begin{figure*}[h!]
    \centering
    \includegraphics[width=.8\linewidth]{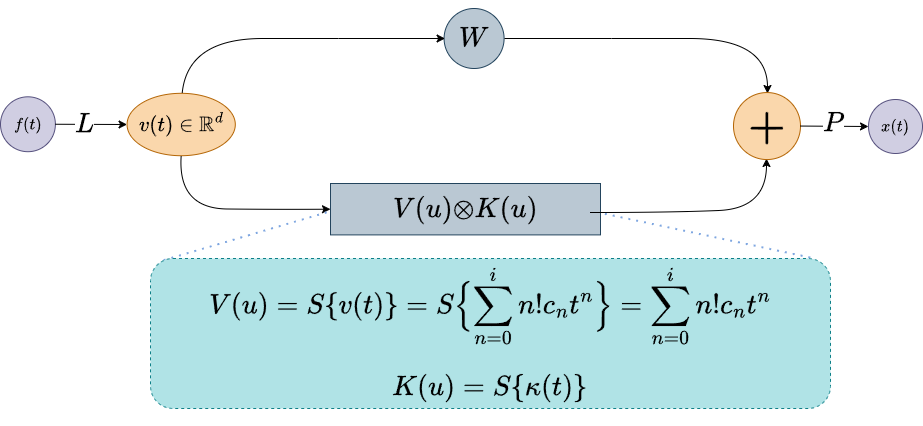}
    \caption{Diagram of the Sumudu Neural Operator (SNO) architecture. Given an input function $f(t)$, we1 lift the input function to a higher dimension using a shallow, linear neural network, $L$, (2) we apply a Sumudu transform, and parametrize weights in the Sumudu space, while applying a local linear transform $W$ as a bias, (3) we normalize results, add our activation function, and project back into the original dimension through a second neural network, $P$, to get our output.}
    
    

    \label{fig:1}
\end{figure*}

To achieve the proposed SNO, we first raise the input function to a higher-dimensional representation using a lifting function, $L$. We realize this through a shallow linear neural network. We then apply a standard neural net operation in which we sum a Sumudu layer, acting as a weight, and a bias function, $W$. 


\paragraph{The Sumudu Neural Operator} To build our neural operator we first start with a general structure for a neural operator,
\begin{equation}
    u(t) = \sigma(W+(\mathcal{K}(\mathbf{f}(t);\phi)(x)); x \in D
\end{equation}
Where $u(t)$ is the high dimensional analogue of the output $x(t)$ which is in domain $D$, $\sigma$ is a nonlinear activation function, $W$ is a weight achieved through a linear transformation, $\mathcal{K}$ is a kernel integral transformation:
\begin{equation}
   \mathcal{K} = \int_{D}\kappa_\phi(t,\tau, \mathbf{f}(t), \mathbf{f}(\tau);\phi)v(\tau)d\tau,
\end{equation}
and $\phi$ is the network parameters. We now replace $\mathcal{K}$ with the Sumudu Transform which is defined as,

\begin{equation}
    S\{f(t)\}(u) = \int_{0}^{\infty} f(ut) e^{-t} \, dt.
\end{equation}
In the next section, we construct a method of calculating the Sumudu transform through a fitted polynomial. We will also describe the time efficiencies this offers.
\paragraph{Calculating the Sumudu Transform with Polynomials}
The Sumudu Transform has a unique property regarding the polynomial expansion of an input signal and its corresponding transformed signal. Specifically, the polynomial regression allows $S\{f(t)\}$ to approximate the polynomial expansion of $f(t)$ up to a factorial scaling factor as shown by \cite{Glen2019Tutorial}:

\begin{equation}
    S\{f(t)\}(u) = \sum_{n=0}^{\infty}n!c_nt^n
\end{equation}
 
Where $c_n$ is the $nth$ coefficient of the polynomial expansion of $f(t)$. This property suggests computational benefits for SNO, particularly for efficient forward and inverse transforms. This can be easily computed using a time complexity comparison; the Cooley-Tukey FFT algorithm has a time complexity of $O(n\log n)$, and polynomial regression has a time complexity of $O(d^2n)$-for degree $d$. Because linear regression with the least-squares method becomes numerically unstable for high degrees, the degree parameter becomes negligible when $n$ is sufficiently high and the degree is low, resulting in complexity scaling as $O(n)$. This demonstrates superior theoretical efficiency relative to FNO and LNO for sufficiently large $n$. Then, a neural operator using the Sumudu transform can be implemented using a similar approach to \citet{cao2023lno}.

We build the polynomial by decomposing the input via polynomial regression with a Vandermonde matrix:
\begin{equation}
    \begin{bmatrix}
        1 & x_0 & \dots& x_0^{d-1} &x_0^d\\
        1 & x_1 & \dots &x_1^{d-1}&x_1^d\\
        \vdots & \vdots & \ddots & \vdots & \vdots\\
        1 & x_{n-1} & \dots & x_{n-1}^{d-1} & x_{n-1}^d\\
        1 & x_n & \dots &x_n^{d-1} &x_n^d\\
    \end{bmatrix}
    \begin{bmatrix}
        a_0\\
        a_1\\
        \vdots\\
        a_{n-1}\\
        a_n\\
    \end{bmatrix}
    =
    \begin{bmatrix}
        y_0\\
        y_1\\
        \vdots\\
        y_{n-1}\\
        y_n\\
    \end{bmatrix}
\end{equation}
Where $d$ is the degree of the regression and $n$ is the number of data points. This is then solved by minimizing the squared error. This is calculated using the Moore-Penrose inverse matrix. This can be done in $O(nd^2)$ time complexity or simply said, it scales with $O(n)$ time complexity. This beats the FFT Cooley-Tukey algorithm which has time complexity of $O(nlog(n))$ \cite{CooleyTukey1965}. This theoretical limit is tested in the results section \ref{tab:runtime} using a runtime analysis in which the programs are run simultaneously to avoid externalities.

\paragraph{Task.} We assess the SNO on problems previously used in Fourier and Laplace Neural Operators, which include ordinary differential equations (ODEs) and partial differential equations (PDEs). We run our benchmark on Burgers’ equation, Duffing oscillators, the Driven Pendulum equation, and the Lorenz system. We utilize the same training and testing datasets as those used in prior work. All input data are first normalized before being input into the system:

\paragraph{Training.} We perform training with the Adam optimizer \cite{KingmaBa2015Adam} using a learning rate of $1 \times 10^{-3}$ and a batch size of 20, for 1000 epochs(Pendulum used 1200 epochs and batch size 50). We use the relative $L^2$ error between the predicted solution and the ground-truth solution as our loss function, as in \cite{li2021fourier}.

\paragraph{Evaluation.} Model accuracy is measured by relative $L_2$ error between predicted and ground-truth solutions on out-of-distribution functions. Performance of the SNO is compared to that of the LNO and FNO on each task. We also compare run-time efficiencies between FNO, LNO, and SNO, using torch synchronization. The times are not wholly representative of actual efficiencies due to differences in code and optimization.

\section{Results}

\begin{figure*}
    \centering
    \includegraphics[width=1.0\linewidth]{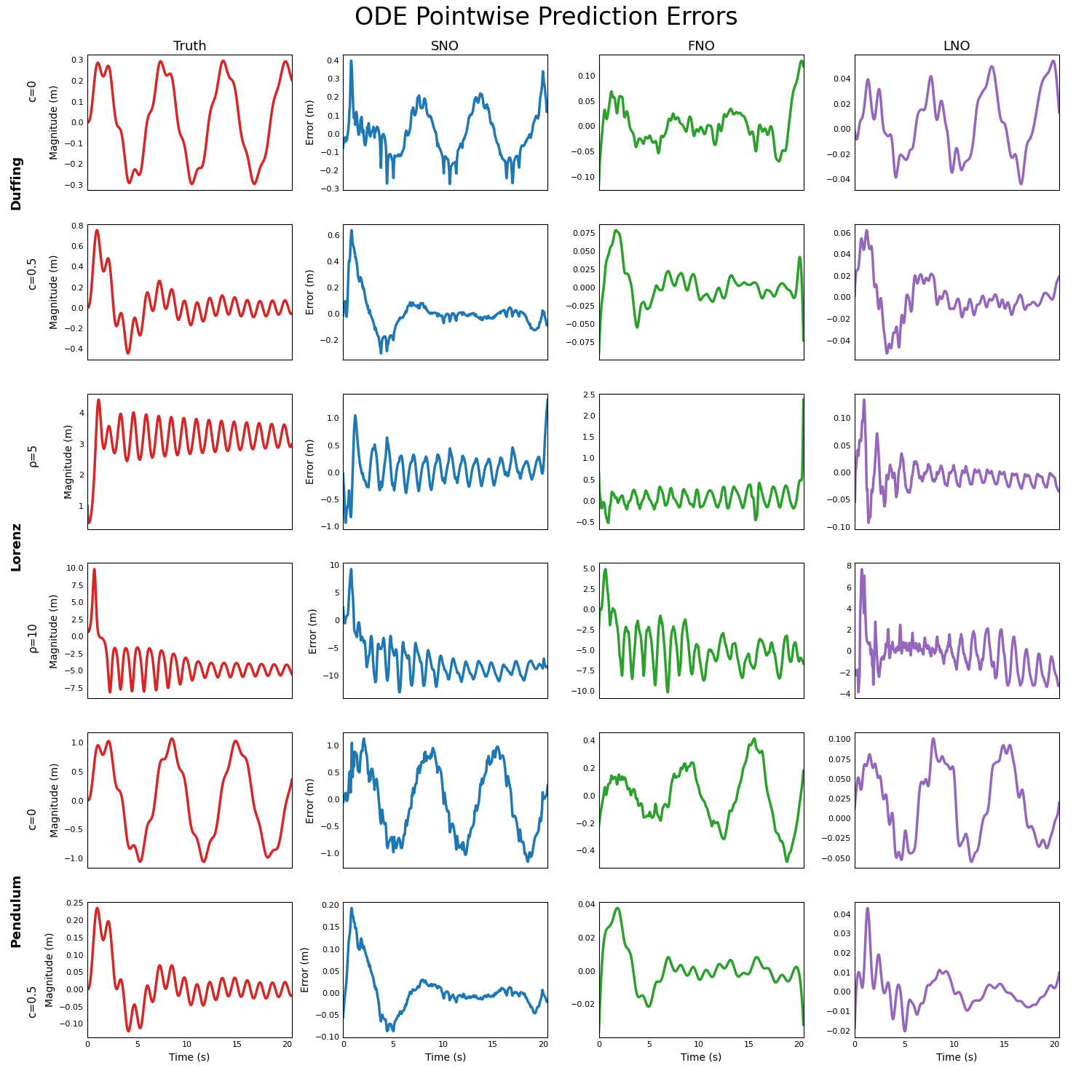}
    \caption{ODE ground truths and pointwise prediction errors between FNO, LNO, and SNO.}
\end{figure*}

\begin{figure*}
    \centering
    \includegraphics[width=1.0\linewidth]{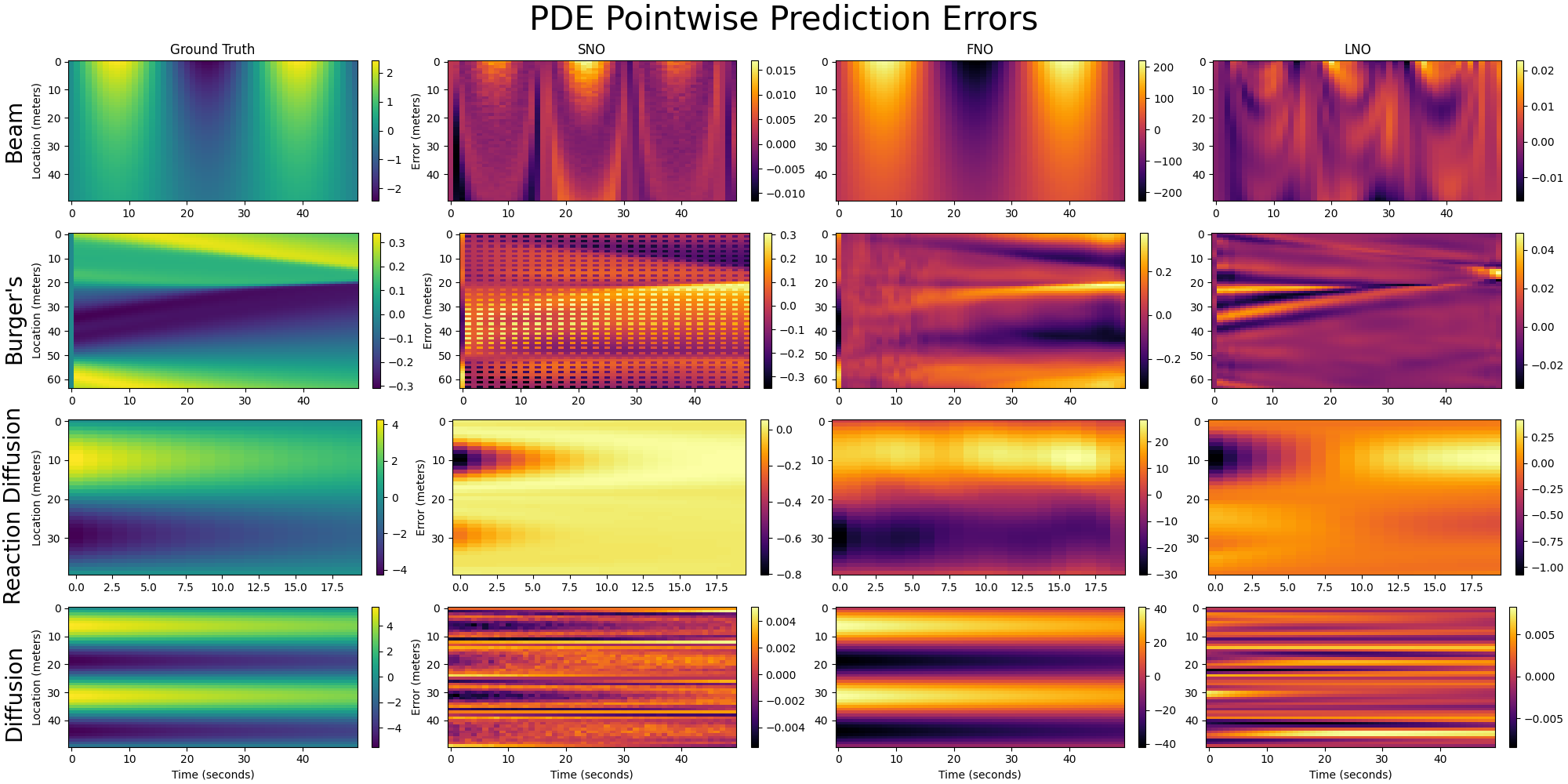}
    \caption{PDE ground truths and pointwise prediction errors between FNO, LNO, and SNO.}
\end{figure*}

\begin{table*}[ht]
\centering
\caption{Average pointwise prediction errors on function types between FNO, LNO, and SNO models.} 
\label{tab:prediction-error}
\begin{tabular}{lcccc}
\toprule
\multicolumn{2}{c}{\textbf{Function Type}} & \textbf{SNO} & \textbf{FNO} & \textbf{LNO} \\
\midrule
\multirow{6}{*}{ODEs} & Duffing Oscillator ($c=0$) & 0.9125 & 0.8761 & \textbf{0.2777} \\
& Duffing Oscillator ($c=0.5$) & 0.7147 & 0.3343 & \textbf{0.1372} \\
& Pendulum ($c=0$) & 0.9100 & 0.5150 & \textbf{0.1961} \\
& Pendulum ($c=0.5$) & 0.7386 & 0.1893 & \textbf{0.1332} \\
& Lorenz System ($\rho=10$) & 0.5565 & 0.4129 & \textbf{0.3024} \\
& Lorenz System ($\rho=5$) & 0.1025 & 0.06110 & \textbf{0.008700} \\
\midrule
\multirow{4}{*}{2D PDEs} & Euler--Bernoulli Beam & \textbf{0.002100} & 1.518 & 0.007900 \\
& Burger & 0.3331 & 3.389 & \textbf{0.05310} \\
& Diffusion & \textbf{0.0006779} & 1.246 & 0.001300 \\
& Diffusion Reaction & 0.1185 & 7.76775652 & \textbf{0.1123} \\
\midrule
3D PDE & Brusselator & \textbf{0.1093} & 0.1283 & 0.1834 \\
\bottomrule
\end{tabular}
\end{table*}

SNO achieves the lowest average error among all three operators on the Euler-Bernoulli Beam (0.0021) and Diffusion equation (0.00068), outperforming even LNO by factors of 3.8× and 1.9× respectively. On Burger's equation, SNO (0.3331) significantly outperforms FNO (3.389) while remaining within an order of magnitude of LNO (0.0531). However, SNO exhibits weaker performance on ODE tasks, particularly for the Duffing Oscillator with $c=0.5$ (0.7147 vs LNO's 0.1372)(\ref{tab:prediction-error}).

We can see that SNO achieves lower error on PDEs compared to ODEs, and furthermore, does better on dampened systems than undampened, visible in the errors.

\subsection{Zero Shot Super Resolution}
We also evaluate SNO's zero-shot super resolution. Using the model weights from the end of training, we evaluate SNO on the test data and plot it against the ground truth and the base resolutions. For the ODEs, we interpolate the ground truth and base resolution to compare against the super resolution. In contrast, for the PDEs, we compare everything at the base resolution to better maintain accuracy. We find that SNO performs better in dampned tasks than in undamped ones, with only slight differences across resolutions. Plots for such comparisons are available in Appendix A.

\section{Discussion}

\begin{table}[t]
\centering
\caption{Runtime comparisons between FNO, LNO, and SNO (seconds).}
\label{tab:runtime}
\begin{tabular}{lccc}
\toprule
\textbf{Problem Type} & \textbf{FNO} & \textbf{LNO} & \textbf{SNO} \\
\midrule
Duffing Oscillator ($c=0$) & 46.33 & 2193.89 & 1913.45 \\
Duffing Oscillator ($c=0.5$) & 49.27 & 3244.41 & 2291.55 \\
Diffusion & 31.47 & 3099.65 & 261.03 \\
Diffusion Reaction & 30.84 & 2532.03 & 84.21 \\
Brusselator & 562.41 & 805.80 & 2241.85 \\
\bottomrule
\end{tabular}
\end{table}

SNO demonstrates competitive and sometimes superior performance compared to established neural operators, particularly on PDE tasks. The strong performance on beam mechanics and diffusion problems suggests that the Sumudu transform's polynomial representation may be especially well-suited for problems with smooth solution manifolds.

The performance gap on ODE tasks (Duffing, Pendulum, Lorenz) indicates that SNO, like FNO, is not built for handling signals that are either undamped or have a transient part or both\footnote{The method of interpolating the polynomial through polynomial regression means sacrificing initial accuracy for general accuracy. In prior experimentation, substituting this with a taylor series approximation centered at any point of importance in the input signal resulted in significantly higher accuracy near that point and less average accuracy which is a potential solution to the transient-state problem with a tradeoff of steady-state accuracy. A mixed interpolation technique may achieve better results on both transient-state and steady-state signal responses.}. This was expected as SNO lacks the pole-residue formulation of LNO, but the performance of SNO on damped tasks surprisingly does not lag to far behind LNO.

Runtime analysis reveals trade-offs: SNO achieves $11.9$ times speedup over LNO on Diffusion and $30$ times speedup on Diffusion-Reaction, though it lags on Brusselator. These results indicate that computational efficiency is problem-dependent and nuanced and warrants deeper analysis of the relationship between problem structure and transform choice. SNO, in general, achieves better runtime efficiencies compared to LNO on most applications. Confounding the experimental runtimes further, the implementations of LNO, SNO, and FNO, vary in actual efficiency. FNO, as a relatively simple implementation, and due to wide research into Fourier Transform, is very fast. LNO has little optimization in its code and is thus not very experimentally efficient. SNO has been optimized, but is still not as efficient as FNO, due to limitations in computation and architecture.

\section{Conclusion}

We introduce the Sumudu Neural Operator (SNO), a novel operator architecture that leverages the Sumudu transform and its connection to polynomial expansions to learn solution operators for ODEs and PDEs more efficiently. Our trials demonstrate that SNO can achieve better performance on most PDEs than both FNO and LNO, despite lagging behind on ODEs. SNO exhibits promise as an alternative neural operator framework, especially on certain PDE types, where it performs significantly better than other models, as well as dampened and steady-state signals. This suggests the need for further testing on these PDE types to understand the benefits SNO provides over other neural operators. Observed limitations in accuracy and behavior highlight the need for further work, including increasing the complexity of the current architecture, more accurate and efficient methods of polynomial regression, and more nuanced time-efficiency analyses. 

\bibliography{aaai2026}

\appendix
\section{Appendix}

\subsection{A: Zero-Shot Super Resolution Comparisons}
\label{app:zero-shot-super-res-plots}

\begin{figure*}[t]
    \centering
    \includegraphics[width=\linewidth]{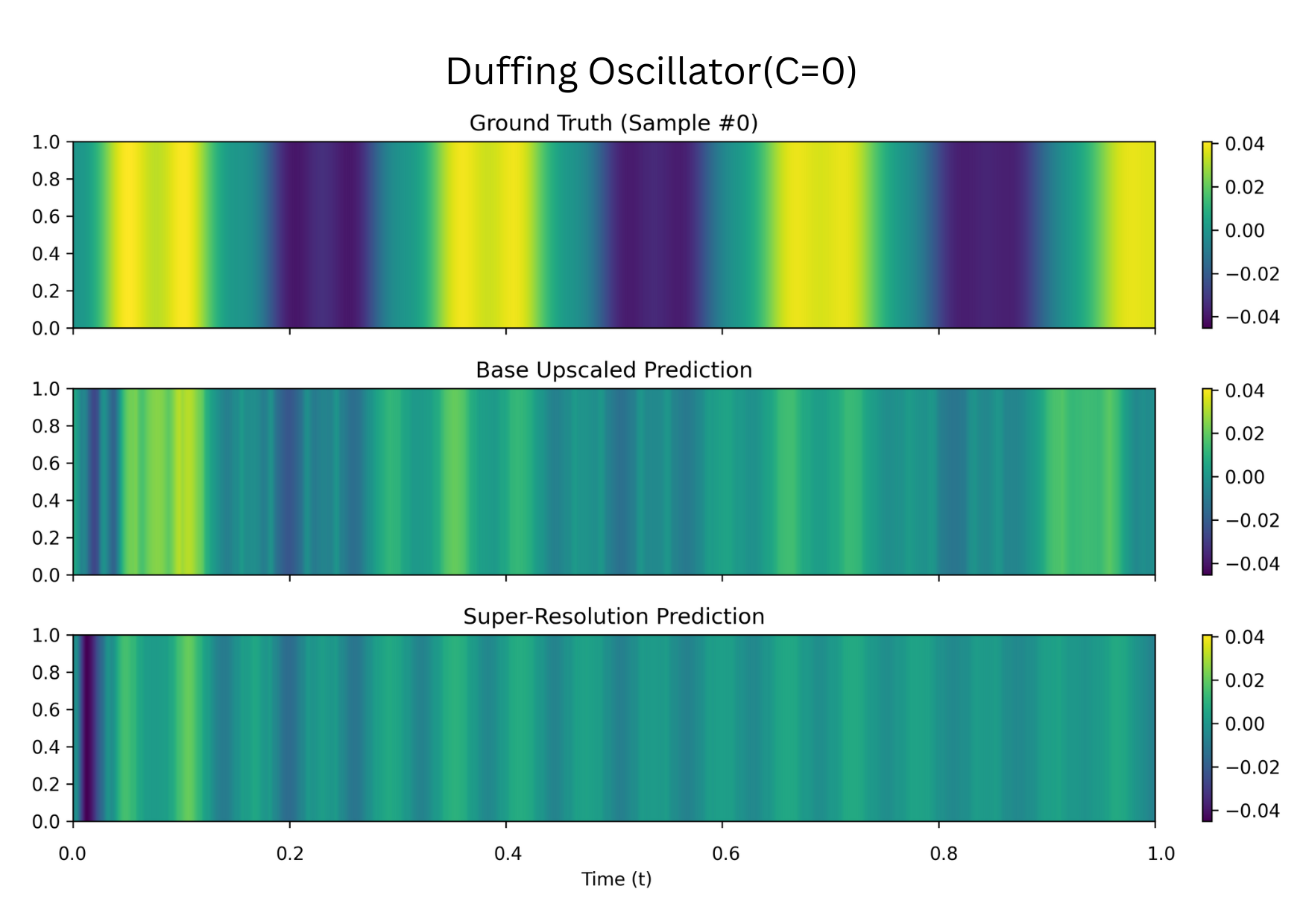}
    \caption{Heatmap comparison of baseline and super-resolved predictions on the Duffing system with zero forcing.}
    \label{fig:duff0}
\end{figure*}

\begin{figure*}[t]
    \centering
    \includegraphics[width=\linewidth]{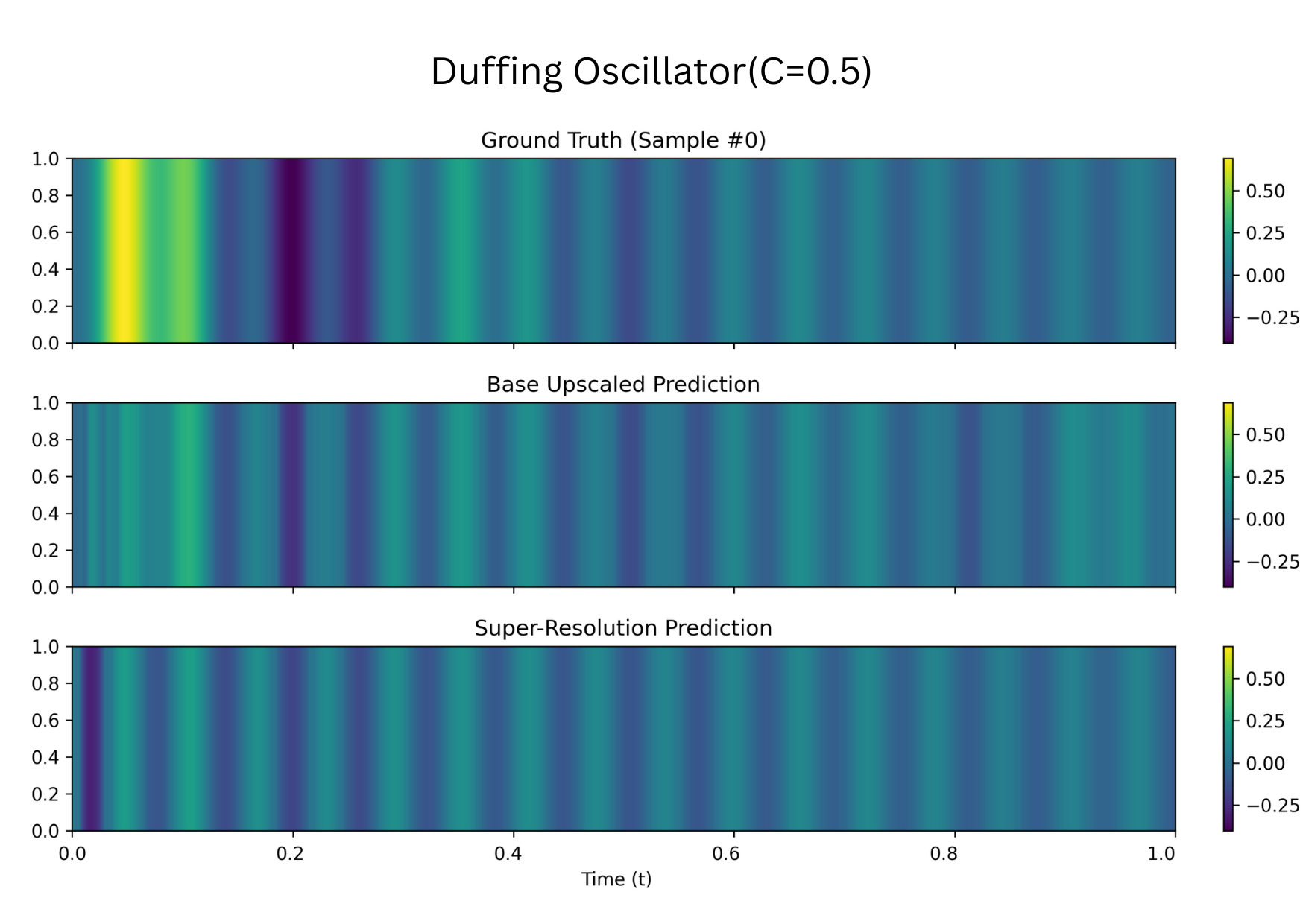}
    \caption{Heatmap comparison of baseline and super-resolved predictions on the Duffing system with moderate forcing.}
    \label{fig:duff5}
\end{figure*}

\begin{figure*}[t]
    \centering
    \includegraphics[width=\linewidth]{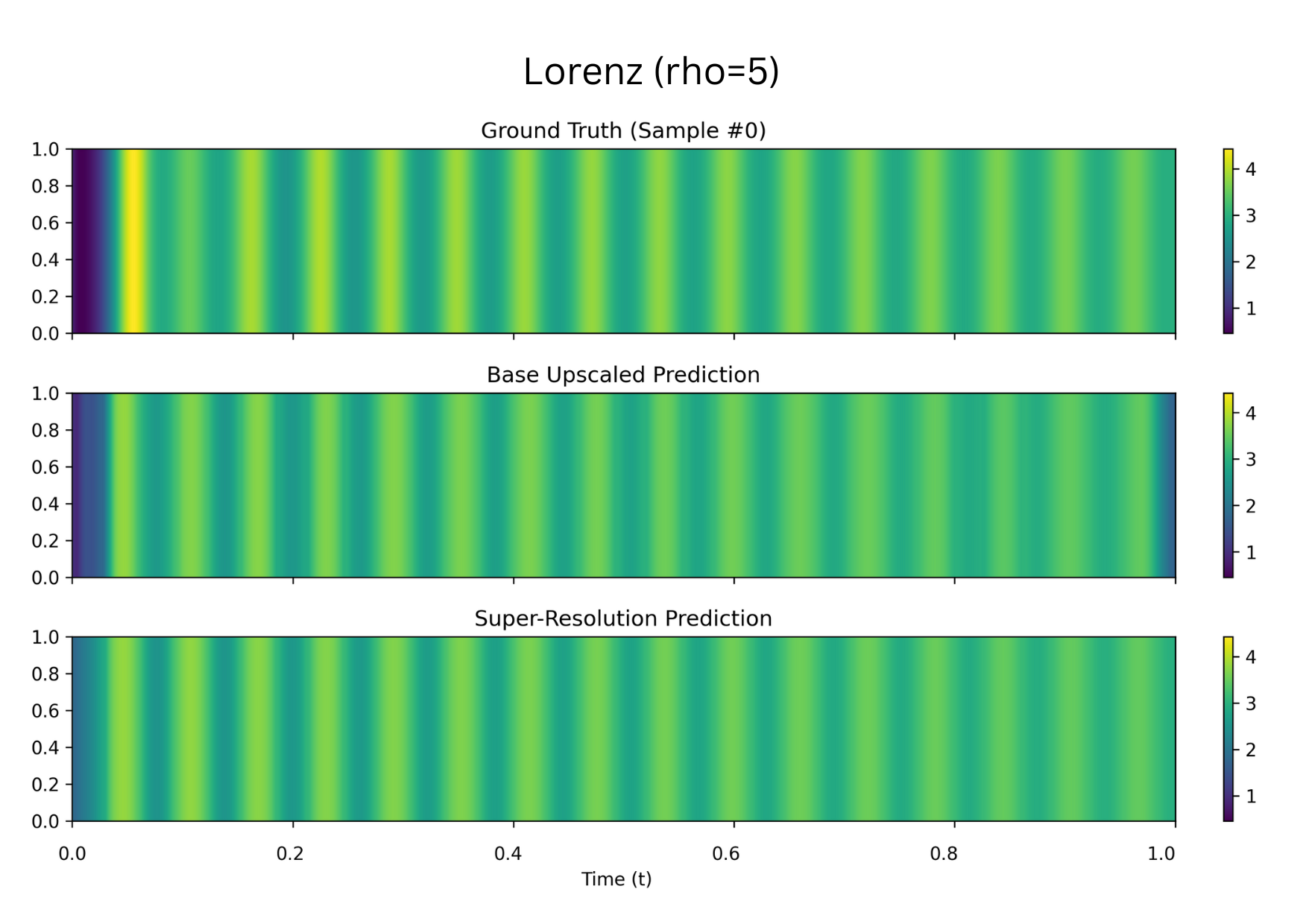}
    \caption{Heatmap comparison of baseline and super-resolved predictions on the Lorenz system with standard parameters.}
    \label{fig:lor5}
\end{figure*}

\begin{figure*}[t]
    \centering
    \includegraphics[width=\linewidth]{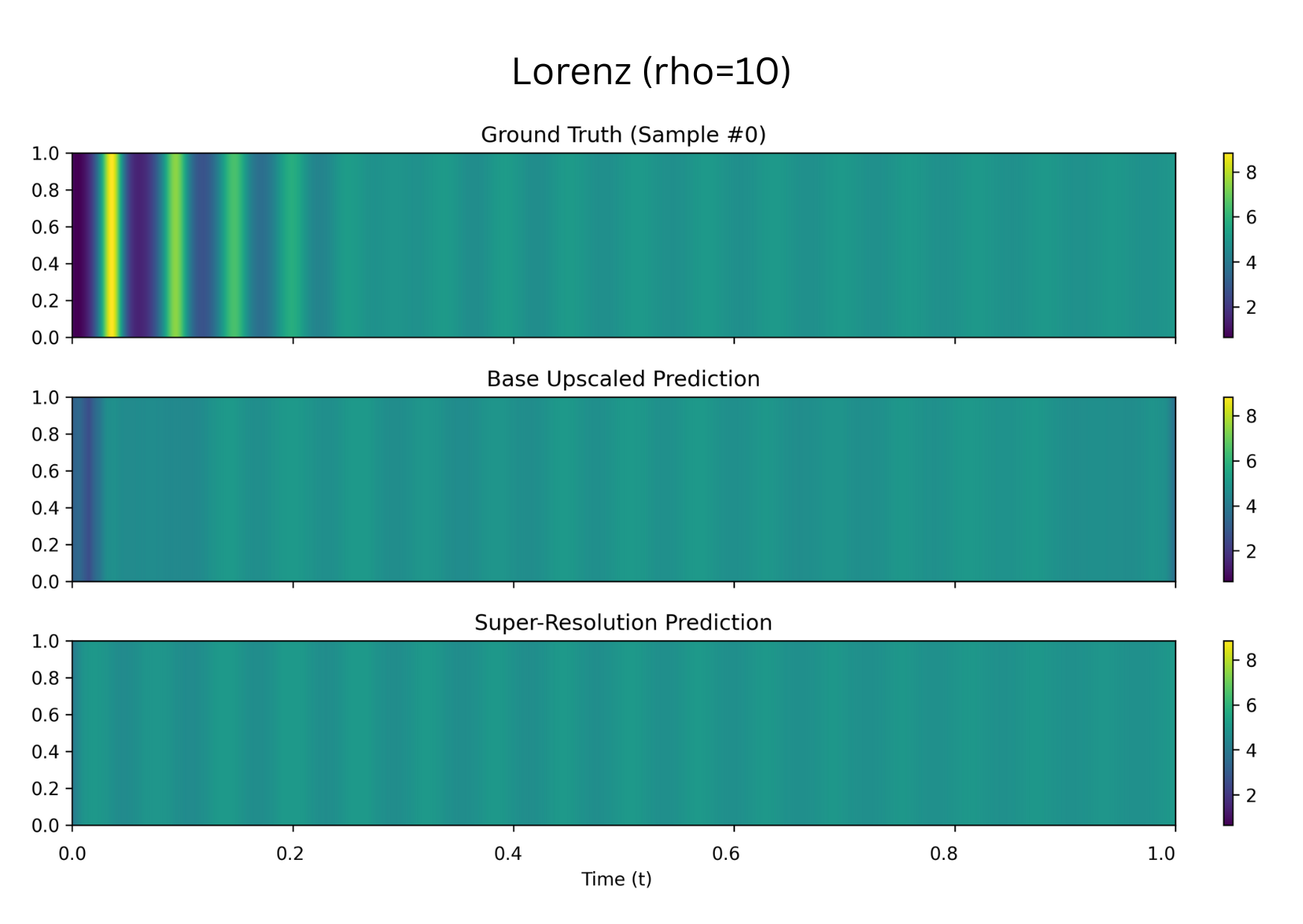}
    \caption{Heatmap comparison of baseline and super-resolved predictions on the Lorenz system with increased chaotic behavior.}
    \label{fig:lor10}
\end{figure*}

\begin{figure*}[t]
    \centering
    \includegraphics[width=\linewidth]{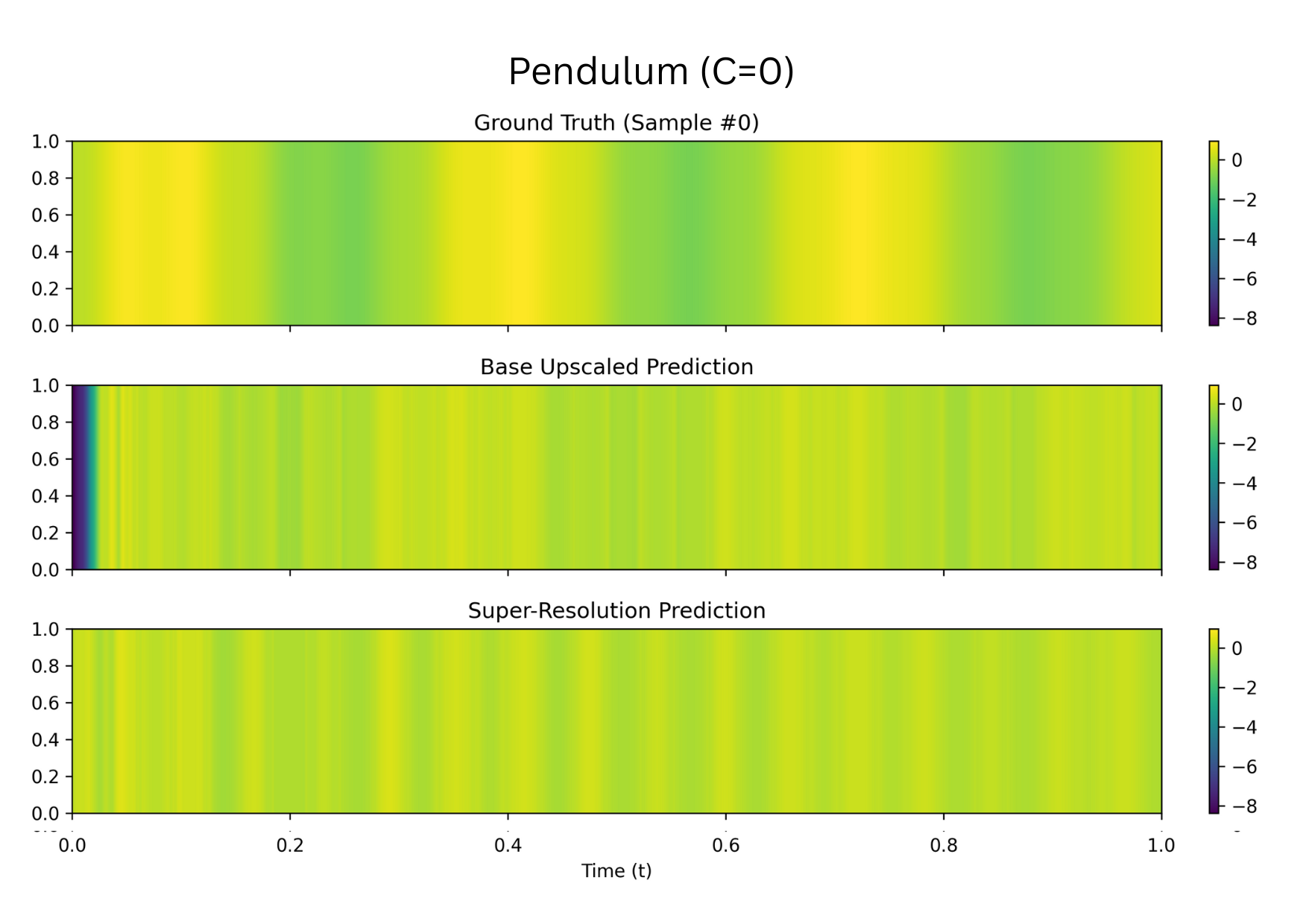}
    \caption{Heatmap comparison of baseline and super-resolved predictions on the pendulum system without external forcing.}
    \label{fig:pen0}
\end{figure*}

\begin{figure*}[t]
    \centering
    \includegraphics[width=\linewidth]{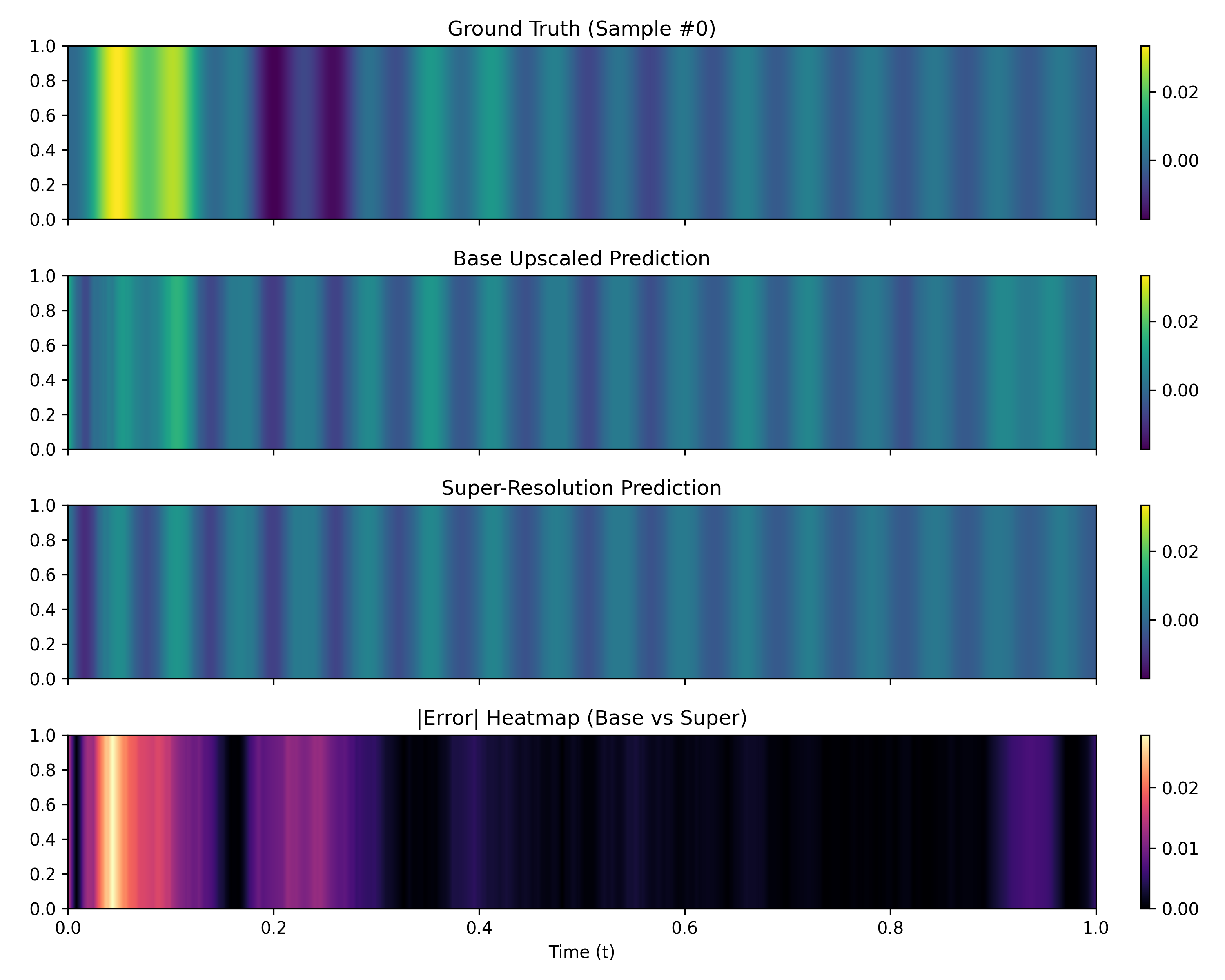}
    \caption{Heatmap comparison of baseline and super-resolved predictions on the pendulum system with moderate external forcing.}
    \label{fig:pen5}
\end{figure*}

\begin{figure*}[t]
    \centering
    \includegraphics[width=\linewidth]{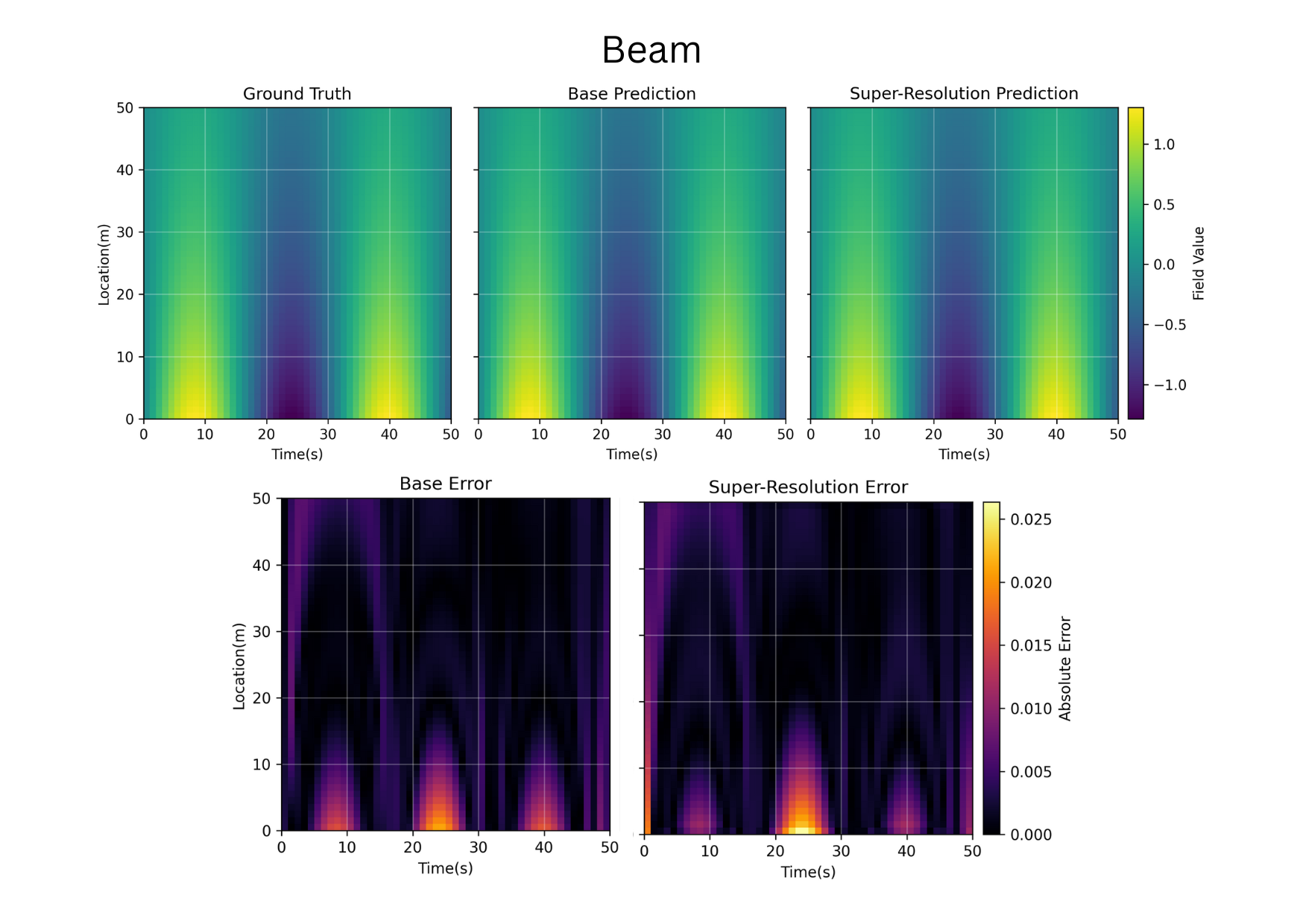}
    \caption{2D heatmap comparison for the elastic beam problem under zero-shot super-resolution.}
    \label{fig:beam_heatmap}
\end{figure*}

\begin{figure*}[t]
    \centering
    \includegraphics[width=\linewidth]{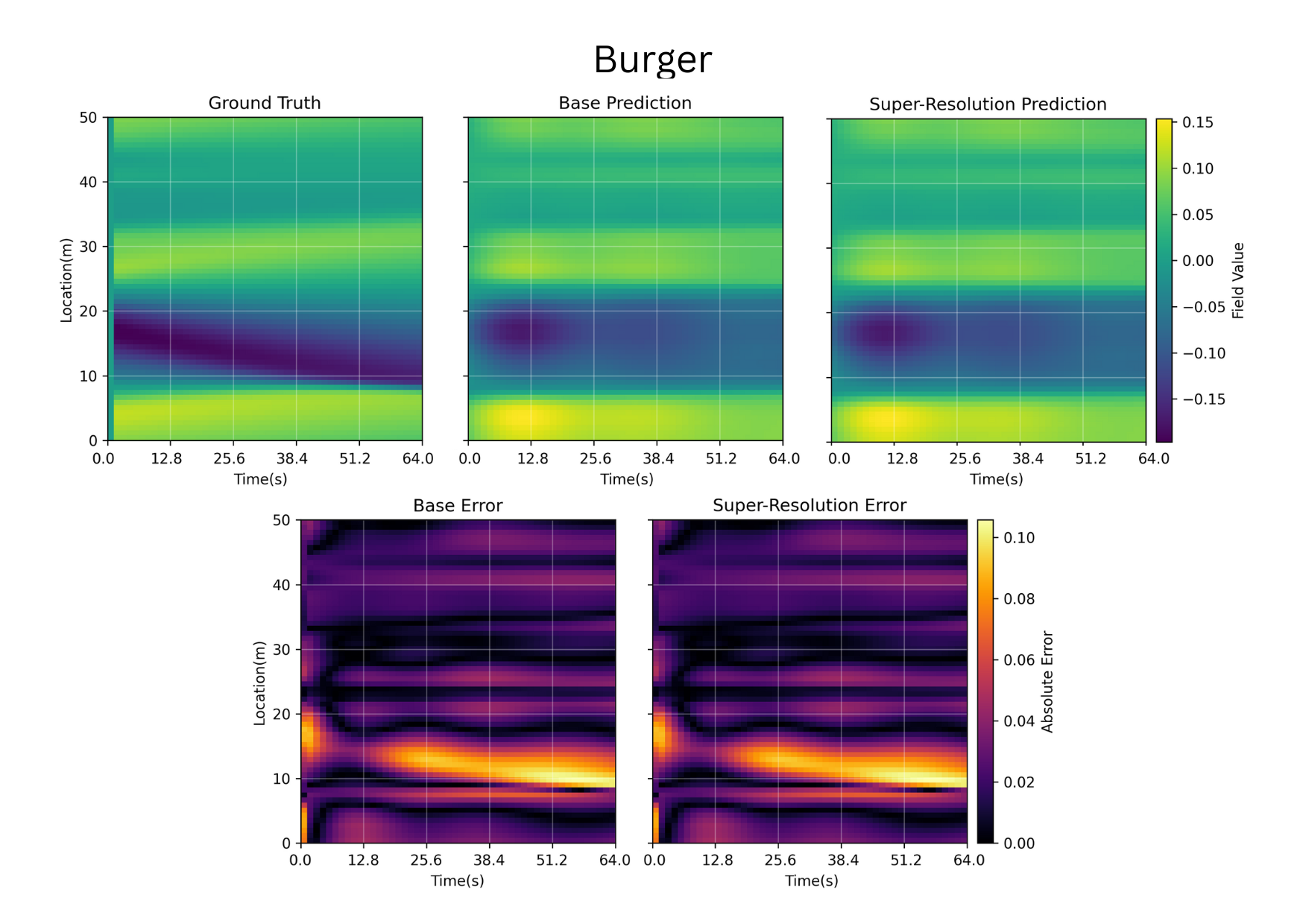}
    \caption{2D heatmap comparison for the Burgers' equation under zero-shot super-resolution.}
    \label{fig:burger}
\end{figure*}

\begin{figure*}[t]
    \centering
    \includegraphics[width=\linewidth]{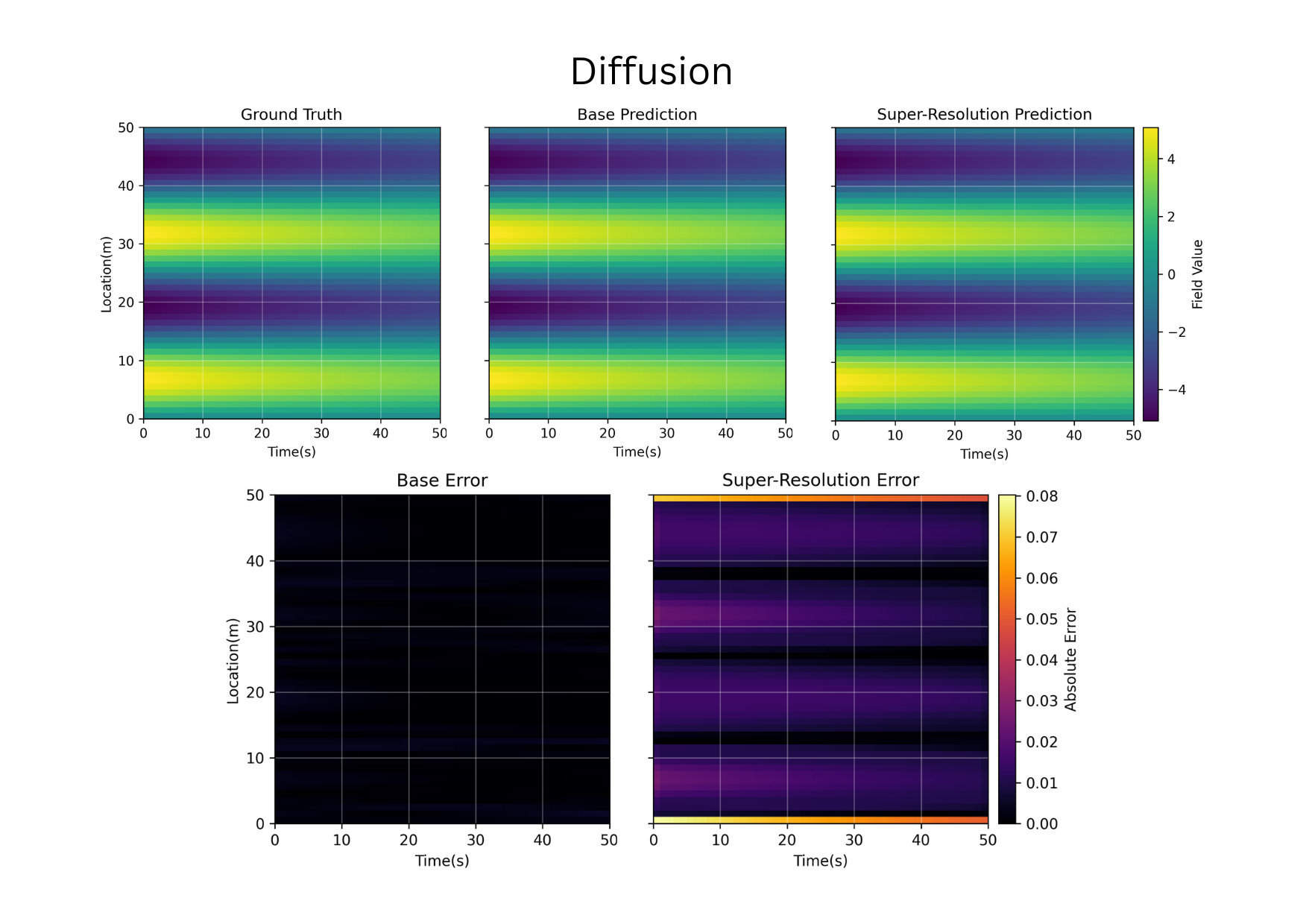}
    \caption{2D heatmap comparison for the diffusion process under zero-shot super-resolution.}
    \label{fig:diffusion_heatmap}
\end{figure*}

\begin{figure*}[t]
    \centering
    \includegraphics[width=\linewidth]{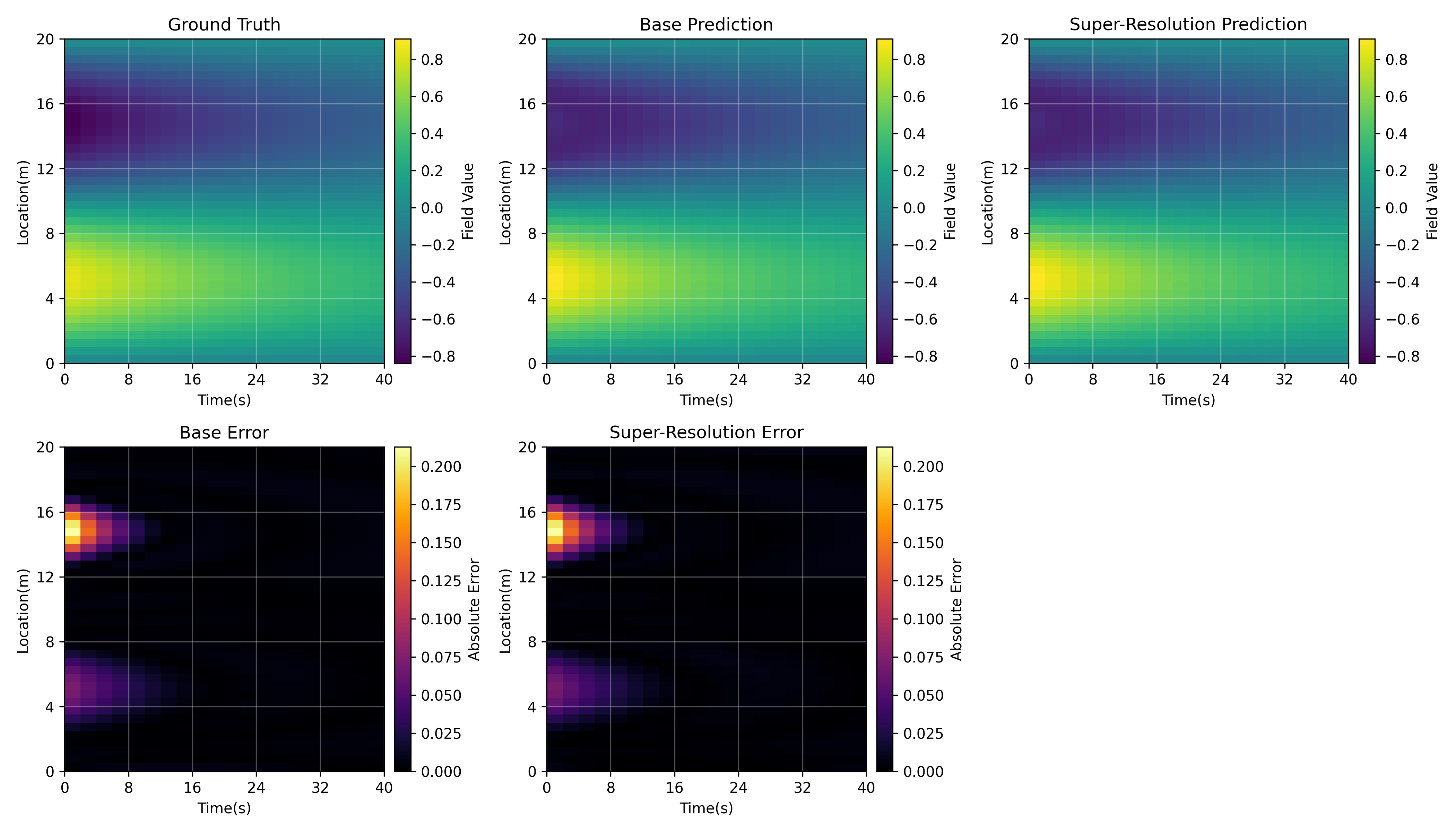}
    \caption{2D heatmap comparison for the reaction–diffusion system under zero-shot super-resolution.}
    \label{fig:rdiff}
\end{figure*}

\subsection{B: Loss Curves}

\begin{figure*}[h!]
    \centering
    \includegraphics[width=\linewidth]{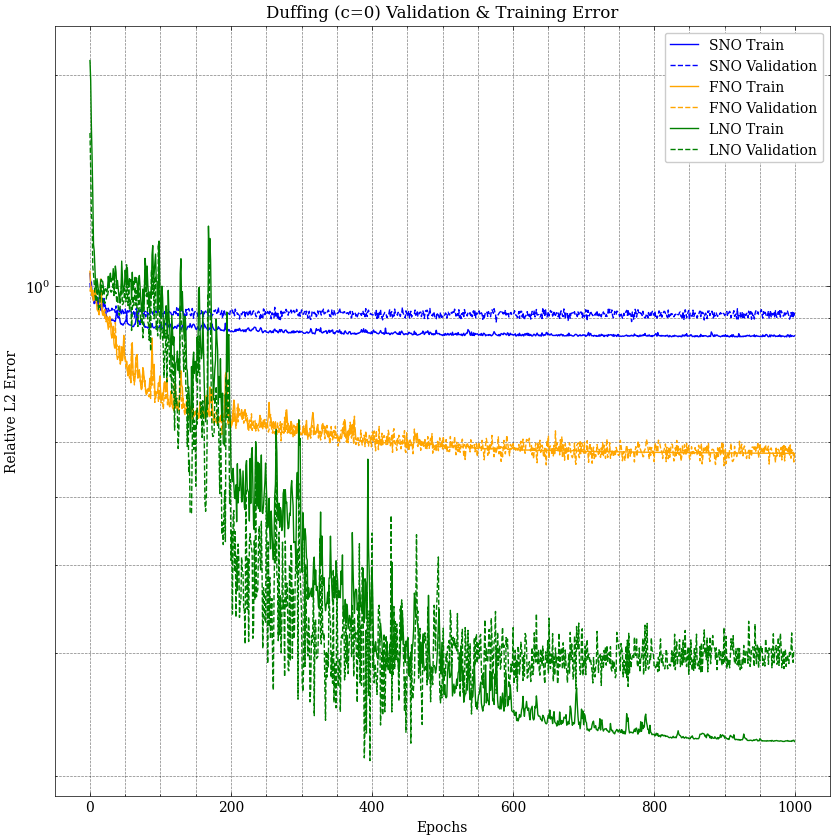}
    \caption{Loss comparison for the Duffing oscillator with damping coefficient \(c = 0\)}
    \label{fig:duffing0}
\end{figure*}

\begin{figure*}[h!]
    \centering
    \includegraphics[width=\linewidth]{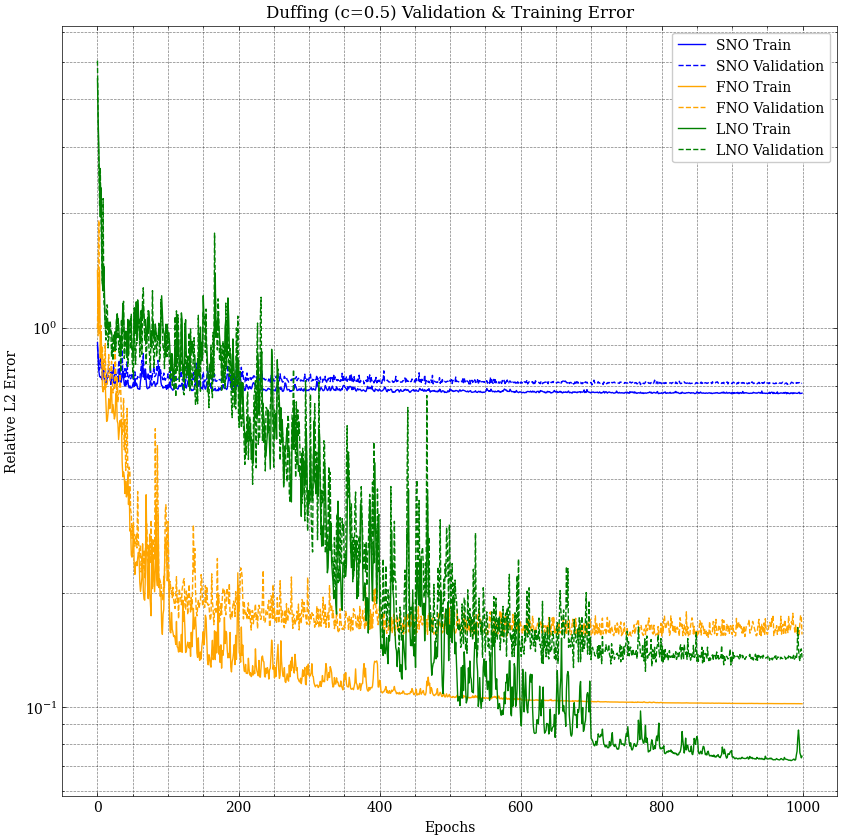}
    \caption{Loss comparison for the Duffing oscillator with damping coefficient \(c = 0.5\)}
    \label{fig:duffing05}
\end{figure*}

\begin{figure*}[h!]
    \centering
    \includegraphics[width=\linewidth]{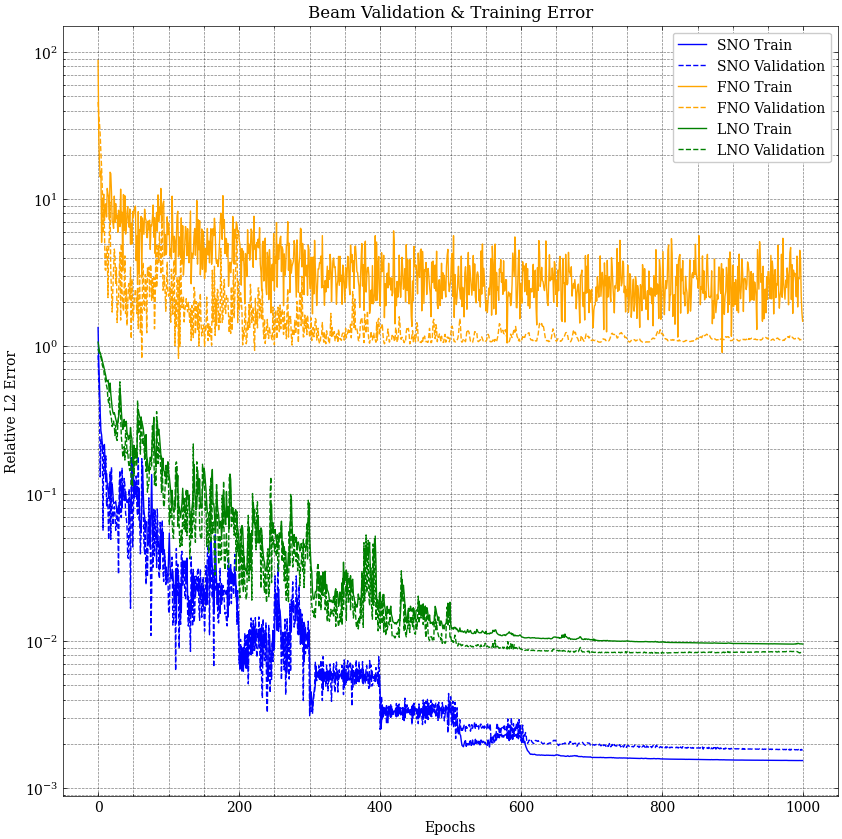}
    \caption{Loss comparison for the beam equation}
    \label{fig:beam}
\end{figure*}

\begin{figure*}[h!]
    \centering
    \includegraphics[width=\linewidth]{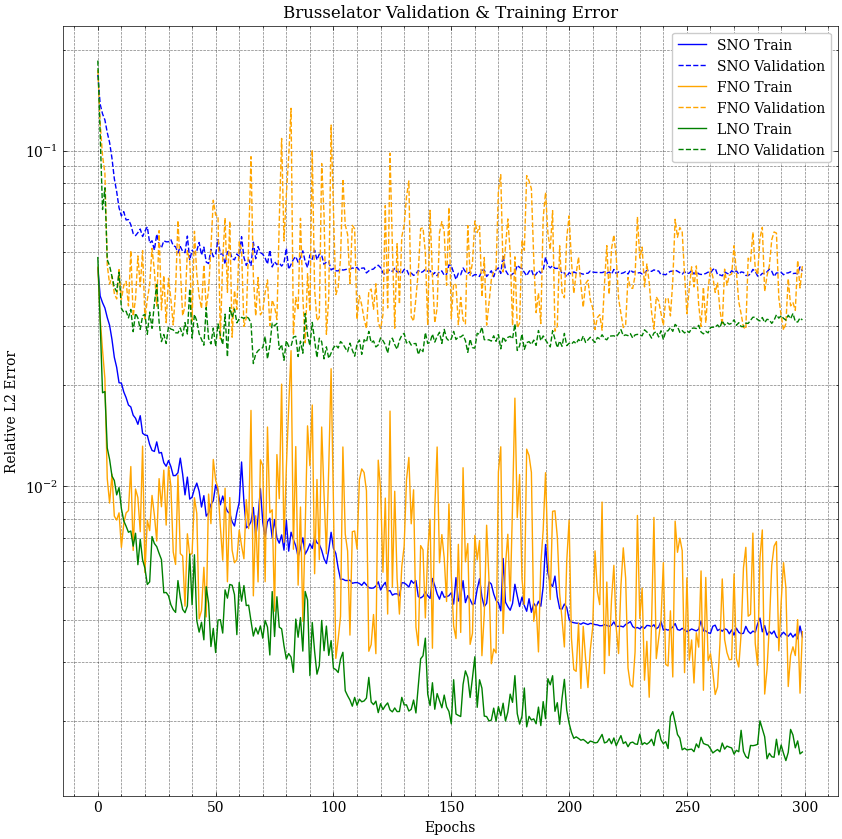}
    \caption{Loss comparison for the Brusselator system}
    \label{fig:brusselator}
\end{figure*}

\begin{figure*}[h!]
    \centering
    \includegraphics[width=\linewidth]{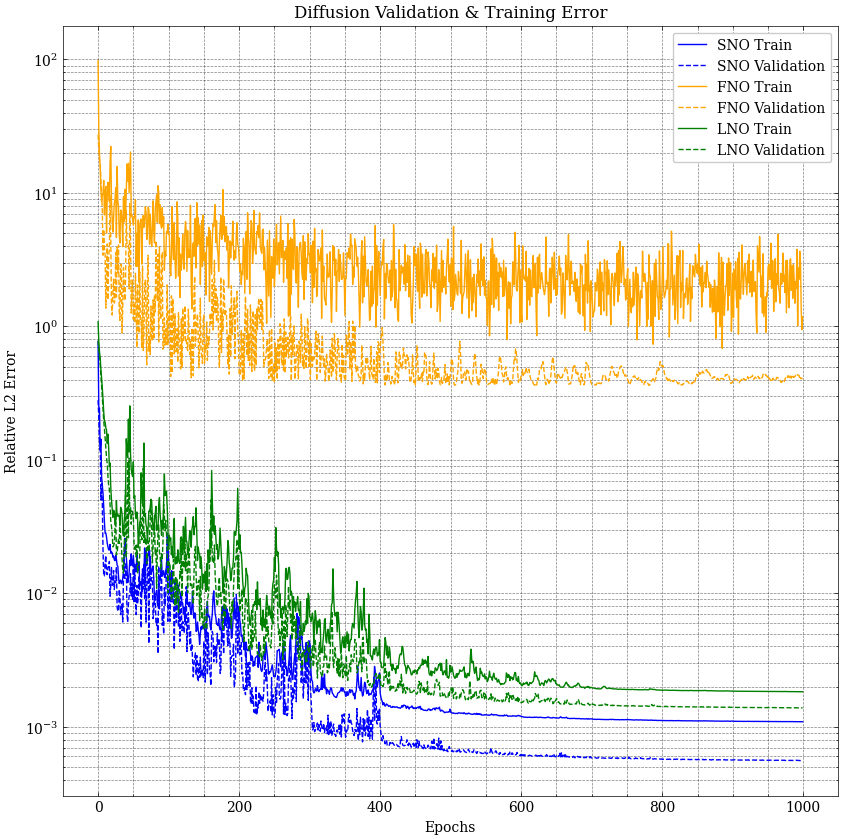}
    \caption{Loss comparison for the diffusion equation}
    \label{fig:diffusion}
\end{figure*}

\begin{figure*}[h!]
    \centering
    \includegraphics[width=\linewidth]{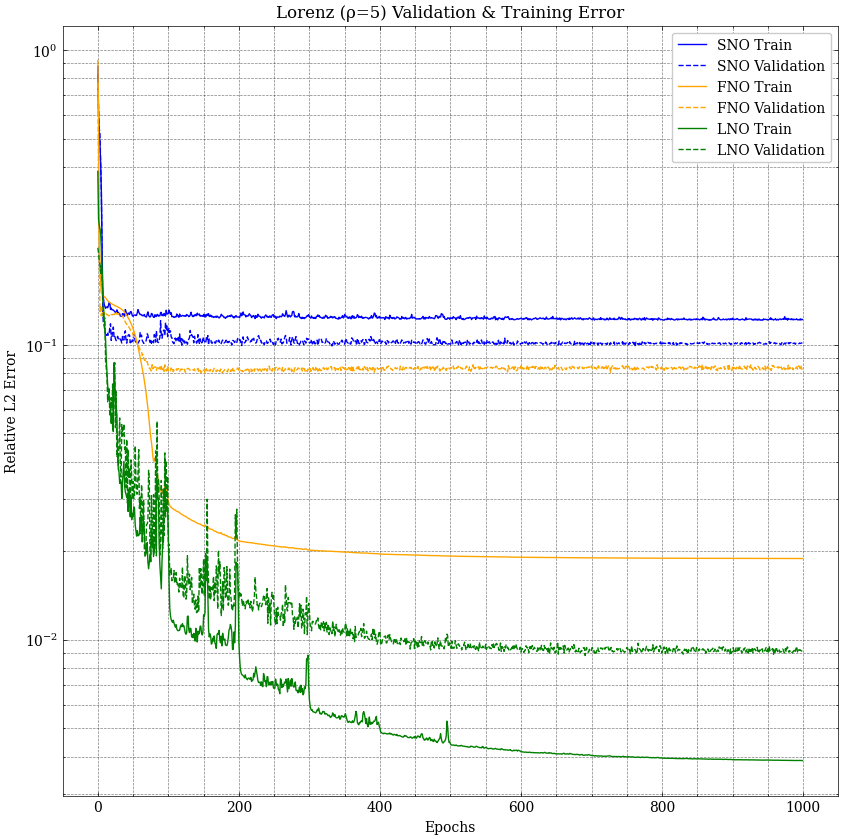}
    \caption{Loss comparison for the Lorenz system with \(\rho = 5\)}
    \label{fig:lorenz5}
\end{figure*}

\begin{figure*}[h!]
    \centering
    \includegraphics[width=\linewidth]{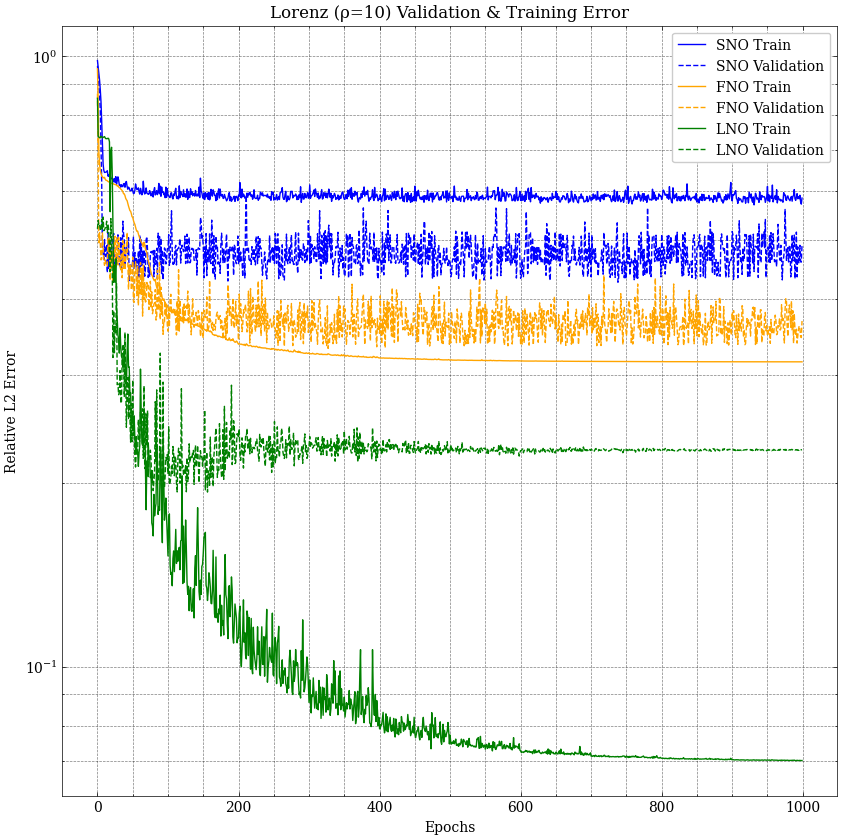}
    \caption{Loss comparison for the Lorenz system with \(\rho = 10\)}
    \label{fig:lorenz10}
\end{figure*}

\begin{figure*}[h!]
    \centering
    \includegraphics[width=\linewidth]{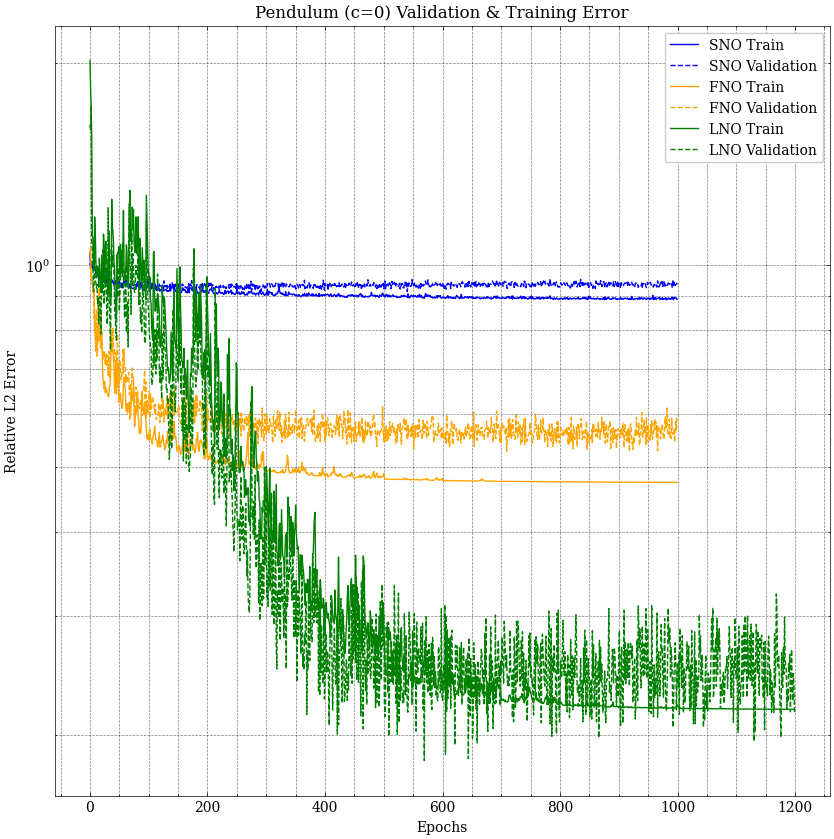}
    \caption{Loss comparison for the pendulum with no damping}
    \label{fig:pendulum0}
\end{figure*}

\begin{figure*}[h!]
    \centering
    \includegraphics[width=\linewidth]{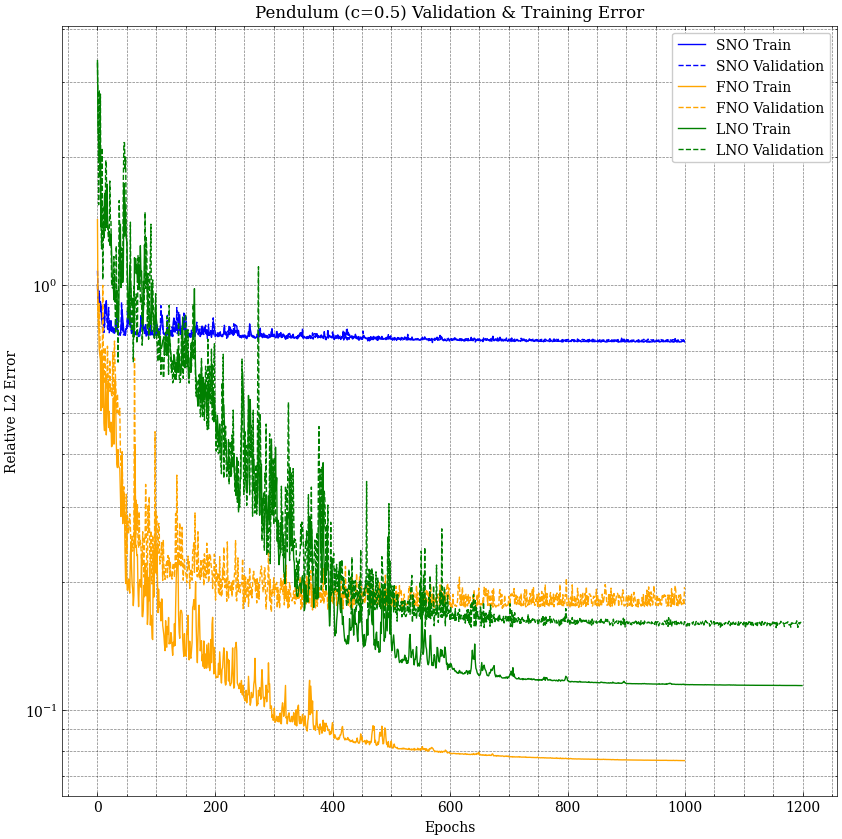}
    \caption{Loss comparison for the pendulum with damping coefficient \(0.5\)}
    \label{fig:pendulum05}
\end{figure*}

\begin{figure*}[h!]
    \centering
    \includegraphics[width=\linewidth]{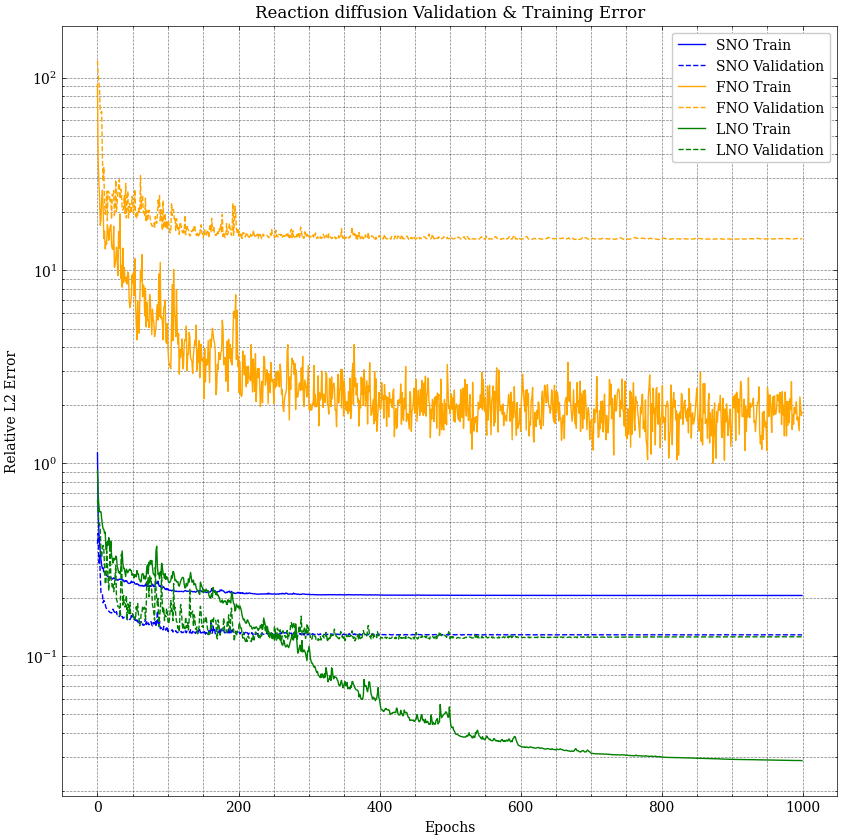}
    \caption{Loss comparison for the reaction--diffusion system}
    \label{fig:reactiondiffusion}
\end{figure*}

\begin{figure*}[h!]
    \centering
    \includegraphics[width=\linewidth]{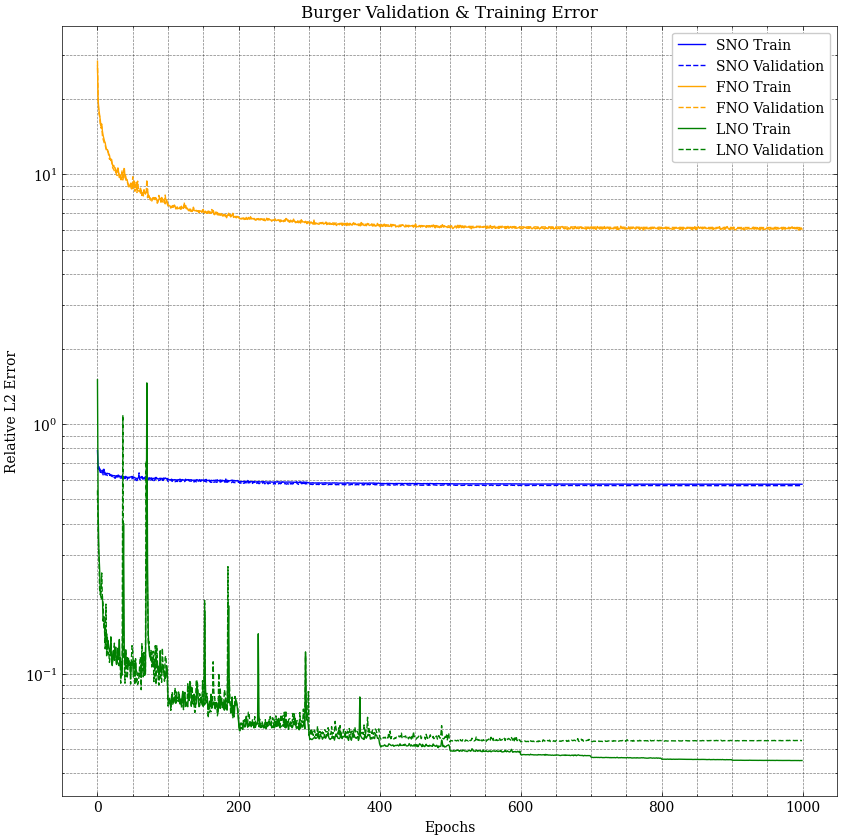}
    \caption{Loss comparison for Burgers' equation}
    \label{fig:burgers}
\end{figure*}


\end{document}